\documentclass{article}

\usepackage[preprint]{neurips_2025}

\usepackage[utf8]{inputenc} % allow utf-8 input
\usepackage[T1]{fontenc}    % use 8-bit T1 fonts
\usepackage{hyperref}       % hyperlinks
\usepackage{url}            % simple URL typesetting
\usepackage{booktabs}       % professional-quality tables
\usepackage{amsfonts}       % blackboard math symbols
\usepackage{nicefrac}       % compact symbols for 1/2, etc.
\usepackage{microtype}      % microtypography
\usepackage{xcolor}         % colors

\usepackage{lipsum}  
\usepackage[linesnumbered,ruled,vlined]{algorithm2e}
\usepackage{amsmath} 
\usepackage{bm}
\usepackage{booktabs}
\usepackage{multirow}
\usepackage{caption} % DO NOT CHANGE THIS AND DO NOT ADD ANY OPTIONS TO IT
\usepackage{times}  % DO NOT CHANGE THIS
\usepackage{helvet}  % DO NOT CHANGE THIS
\usepackage{courier}  % DO NOT CHANGE THIS
\usepackage{graphicx} % DO NOT CHANGE THIS
\usepackage{wrapfig}
\usepackage{caption}

\title{Vocabulary Customization for Efficient Domain‑Specific LLM Deployment}

\author{
  \textbf{Christian Herold}\thanks{Corresponding author. Email: cherold@ebay.com.} \quad \textbf{Michael Kozielski} \quad \textbf{Nicholas Santavas} \\  \textbf{Yannick Versley} \quad  
  \textbf{Shahram Khadivi}\\ 
  eBay Inc.}

\begin{document}

\maketitle

\begin{abstract}

When using an LLM to process text outside the training domain(s), an often
overlooked factor is vocabulary mismatch, where the general-domain tokenizer
fails to capture frequent domain-specific terms, leading to higher token fertility
and thus a decrease in processing speed due to suboptimal sub-word splits.

We address this limitation by augmenting the pretrained vocabulary with a set of domain‑specific tokens. To this end, we design an algorithm that extends an existing tokenizer while guaranteeing it never decreases tokenization efficiency: every input sequence is segmented into at most the same number of tokens as before.

Evaluated on real‑world e‑commerce use-cases, the augmented tokenizer signi\-ficantly shortens input sequences by up to 20\% and reduces inference latency on downstream tasks while preserving predictive quality. We further analyze secondary effects, such as the impact on forward pass speed and the rate at which the model adopts the newly introduced tokens, to illustrate the broader benefits of vocabulary adaptation.

\end{abstract}

\section{Introduction}

Large Language Models (LLMs) have been established as state-of-the-art approaches for countless downstream applications across many domains and languages, still it is a common occurrence
that they are adapted either to underrepresented/unseen languages or to new domains,
such as healthcare, finance or e-commerce \citep{artetxe-etal-2020-cross,DBLP:journals/corr/abs-2402-08831,DBLP:journals/corr/abs-2408-12779}. While it is relatively common to see
tokenizer adaptation for new languages (see also Section \ref{sec:related_work}), we argue that extending the tokenizer
can also show significant benefits for domain adaptation. For example, in e-commerce we can improve the modeling of the language
occurring in brand names, stock-keeping units or multilingual descriptors, which
occur with high regularity but are often not represented as single tokens in the
original vocabulary.

When we extend the vocabulary with a tailored set of additional tokens that cover frequent or semantically critical domain terms, there are a number of
tradeoffs and non-trivial questions on which we want to shed light in this paper,
specifically:
(i) How and to what extent should we select candidate tokens so that the resulting token set maximizes compression without bloating the embedding table? 
(ii) Can we modify an existing tokenizer incrementally and deterministically, such that the new segmentation is never worse than the original, thus preserving backward compatibility with legacy inputs? 
(iii) What are the measurable gains, not only in static token counts but also in end‑to‑end inference time and prediction quality, when the augmented tokenizer is paired with a domain-specific model?

\begin{wrapfigure}{r}{0.45\textwidth}
  \centering
  \includegraphics[width=\linewidth]{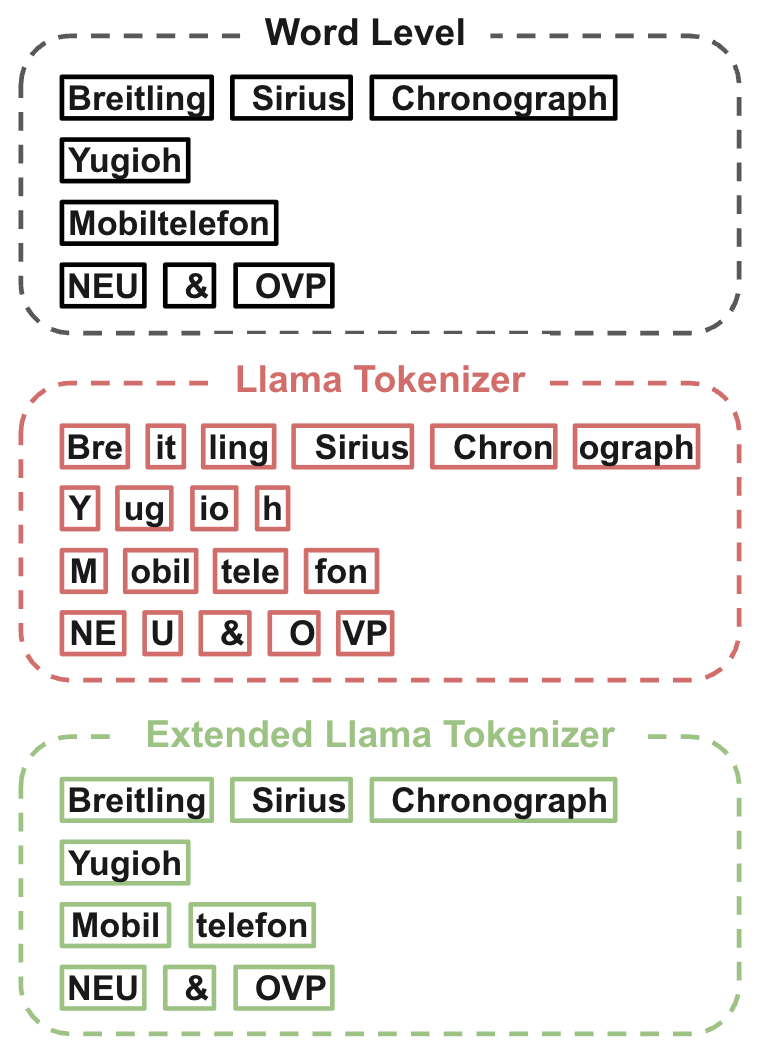} % Reduce the figure size so that it is slightly narrower than the column. Don't use precise values for figure width.This setup will avoid overfull boxes.
\caption{We extend the Llama 3.1 tokenizer with new vocabulary entries and merge operations, which are specific to the e-commerce domain. The result is a much more efficient tokenization for e-commerce specific phrases, which significantly reduces the cost of running such models in production.}
%  \label{fig:intro}
\end{wrapfigure}
We answer these questions through three main contributions. 
\begin{enumerate}
    \item We propose a new algorithm for vocabulary extension that is guaranteed to always result in equal or fewer tokens being used. %\footnote{Code will be published}
    \item We show that vocabulary extension can be applied to make inference with autoregressive LLMs more efficient for a specific domain (e-commerce) by up to 20\% without degrading model quality.
    \item We highlight and investigate multiple implications of vocabulary extensions that are often ignored in previous work, such as the impact on forward pass speed and how frequently the final model utilizes the additional tokens.
\end{enumerate}

By systematically addressing vocabulary mismatch, our work closes an often overlooked gap between laboratory‑grade fine‑tuning and production‑ready LLM deployment.
The proposed method is orthogonal to ongoing efforts in model quantization, speculative decoding, and optimized kernels, and can be combined with them to yield multiplicative efficiency gains, all while maintaining, or even enhancing, application‑level performance.

\section{Related Work}
\label{sec:related_work}

\subsubsection*{Vocabulary Extensions}
Before the rise of autoregressive LLMs, several works have shown that the vocabulary of BERT-style non-autoregressive (encoder-only) models can be extended to adapt the model to a new domain like biomedicine \citep{DBLP:conf/emnlp/Tai0DCK20, DBLP:conf/emnlp/PornerWS20a, DBLP:conf/emnlp/SachidanandaKL21}, news \citep{DBLP:journals/ai/MosinSKTY23}, legal \cite{DBLP:conf/emnlp/GeeZRT22}, and IT \citep{DBLP:conf/emnlp/ZhangRSCFFKRSW20, DBLP:conf/acl/YaoHWDW21}. 
In contrast to the present work, the main motivation to adapt the vocabulary of encoder-only models was to improve the model quality in the new domain, while here we are primarily focused on improving the inference efficiency.
For encoder-only models, the task is also easier because we do not have to worry about if and how well the model decides to produce the new tokens, because these models are not used for text generation, but for classification etc.

In the realm of autoregressive LLMs, previous work is mostly limited to vocabulary extensions for the sake of expanding to new languages.

There exist many approaches that change the existing vocabulary of LLMs to new languages.
This is achieved either by adding new tokens to the existing tokenizer \citep{DBLP:conf/coling/0001XCWLW022, DBLP:journals/corr/abs-2304-08177, DBLP:journals/corr/abs-2311-05845, DBLP:journals/corr/abs-2308-11878, DBLP:journals/corr/abs-2312-13951, DBLP:journals/corr/abs-2401-13303, DBLP:conf/coling/ShaZFS25, DBLP:journals/corr/abs-2404-17790, DBLP:conf/coling/ChoiJPWLKKYPLLH24, DBLP:journals/corr/abs-2312-00738, DBLP:conf/emnlp/TejaswiGC24, DBLP:journals/corr/abs-2407-05841, yamaguchi2024can, yamaguchi2024elchat, DBLP:conf/wmt/LiaoHKM24} or by replacing a certain number of tokens or even the full vocabulary \citep{DBLP:journals/corr/abs-2311-05741, DBLP:journals/corr/abs-2301-09626, DBLP:conf/coling/DaltLBPXGV24, DBLP:journals/corr/abs-2408-04303, DBLP:conf/emnlp/YamaguchiVA24, DBLP:journals/corr/abs-2408-15793, DBLP:conf/emnlp/AlexandrovRMZVT24, DBLP:journals/corr/abs-2410-04335}.
From the above, only very few share the details on how the tokenizer extension is performed.
\citet{DBLP:journals/corr/abs-2311-05741} replace the least frequently used tokens with new ones and add the corresponding merges at the beginning of the merge list.
\citet{DBLP:journals/corr/abs-2312-00738} add new tokens from an existing tokenizer based on some language-specific dataset but do not mention if and how they handle new merge operations.
\citet{yamaguchi2024can} extend the existing tokenizer with new, language-specific tokens by adding the corresponding merges at the beginning of the merge list.

Compared to language-adaptation, there are only few works that investigate vocabulary customization of autoregressive LLMs for domain-adaptation. 
\citet{DBLP:journals/corr/abs-2503-19693} introduce a framework called \textit{AdaptiVocab} where they replace existing tokens with domain-specific n-gram-based tokens to improve decoding efficiency for specific domains.
The downside of their approach is that inputs that do not fall into the right distribution, it can happen that more tokens are needed than with the original tokenization, while our approach guarantees that always equal or fewer tokens are being used.
At the time of writing, they have not yet released their code.
\citet{DBLP:conf/emnlp/LiuWQKKS024} propose an algorithm called \textit{VEGAD} to extend the tokenizer for the legal and medical domains. However, they do not discuss how merge operations are added and they focus on quality improvement rather then efficiency.
\citet{DBLP:conf/icml/DaganSR24} show that inference efficiency on code benchmarks can be improved by adding domain-specific tokens to the model vocabulary. As above work, they also do not discuss how merge operations are added to the existing tokenizer.

Regarding the initialization of the new embedding vectors, previous work has shown that using the average of the existing token embeddings for the respective new tokens performs better than random initialization \citep{DBLP:conf/acl/YaoHWDW21}.

\section{Methodology}
In a subword tokenizer \citep{sennrich-etal-2016-neural},
the list of words is split into pieces using two main components:\footnote{Most tokenizers have further components, like pre- and post-processing operations, special tokens, special templates etc. But these remain unaffected by the tokenizer extension framework and will just be copied over from the original tokenizer.} 
\begin{enumerate}
    \item The \textbf{list of merge operations} tells us which tokens should be merged. A merge operation is a tuple consisting of two strings, e.g., \texttt{(\lq{}th\rq{}, \lq{}e\rq{})}. This example tells us that we need to merge the tokens \texttt{\lq{}th\rq{}} and \texttt{\lq{}e\rq{}}, forming a new token \texttt{\lq{}the\rq{}}. The ordering of the list of merge operations tells us in which order we need to apply these merges.
    \item The \textbf{model vocabulary} is a dictionary that maps each token to a certain index. One entry in the vocabulary could look like this: \texttt{\lq{}th\rq{}: 873}, indicating that the token \texttt{\lq{}th\rq{}} is mapped to the integer \texttt{873}. The value of the index corresponds to the order in which the tokens were added to the tokenizer. That means for example the tokens \texttt{\lq{}t\rq{}} and \texttt{\lq{}h\rq{}} will have an index that is smaller than the index of the token \texttt{\lq{}th\rq{}} which is the result of merging the two.
\end{enumerate}
When the tokenizer is applied to input text, it first decomposes the text at the byte level (for UTF-8 in Western languages, characters).
We then consider the ordered list of merge operations:
For each merge operation, we match the corresponding pairs in the text and merge the two tokens to form a single new token.
After exhausting the full list of merge operations, the model vocabulary maps each remaining token to an integer, which form the LLM's input.

To extend the vocabulary of an existing tokenizer, it is necessary to modify both the model vocabulary and the list of merge operations. In our approach to creating a
domain-adapted tokenizer we have the following steps, described in detail in Appendix \ref{sec:detailed_,ethodology}:
\begin{enumerate}
    \item Training of an in-domain tokenizer using a domain-specific dataset.
    \item Extension of the existing \lq{}original\rq{} tokenizer with new in-domain tokens from the tokenizer trained in (1).
    \item Initialization of the new embedding and projection vectors in the tokenizer-extended LLM.
    \item Continuous training of the tokenizer-extended LLM to optimize for the new vocabulary.
    \item Evaluation of the final tokenizer-extended LLM.
\end{enumerate}

\section{Experiments}

\begin{figure*}[t!]
\centering
  \includegraphics[width=0.40\linewidth]{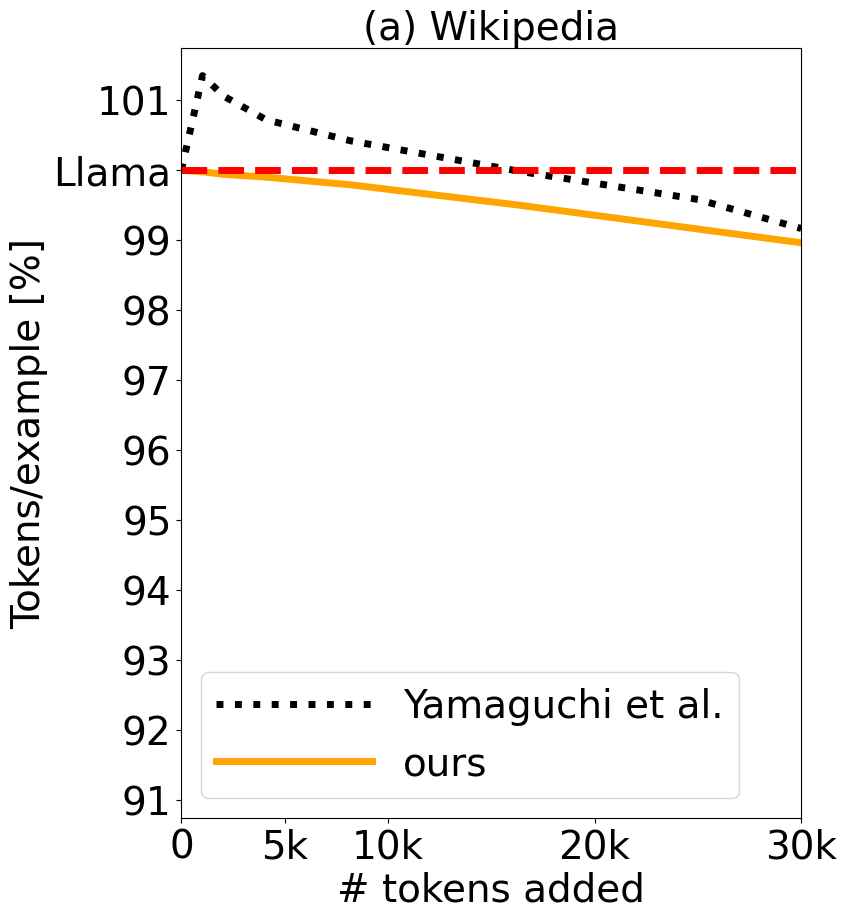} \hspace{1cm}
  \includegraphics[width=0.38\linewidth]{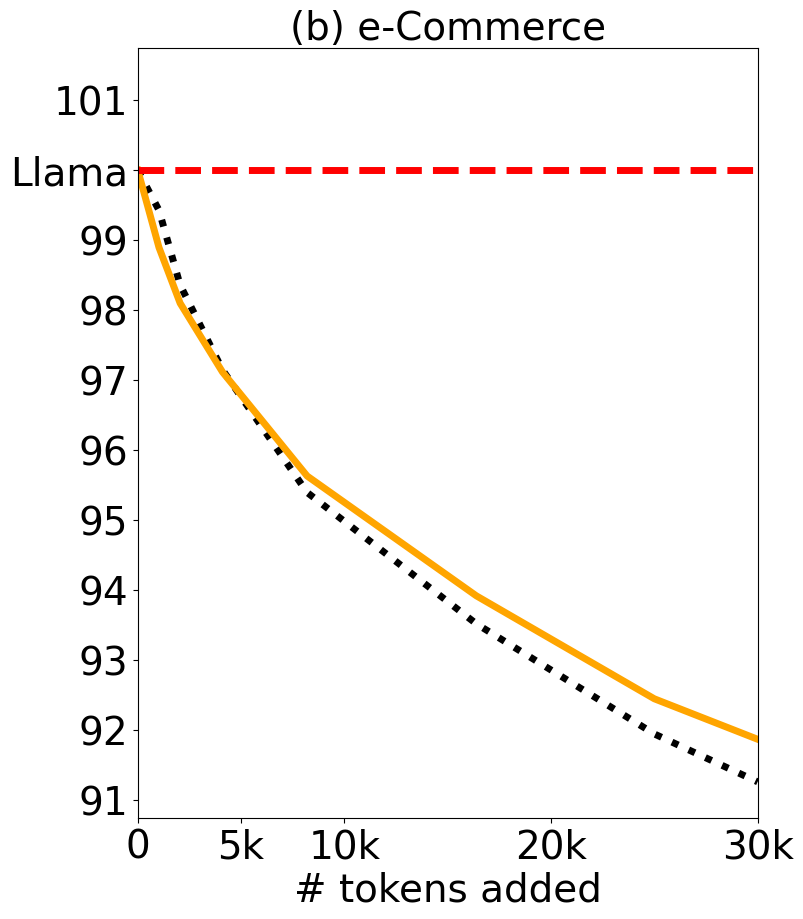}
  \caption {Impact of extending the Llama-3.1 tokenizer with e-commerce specific tokens. Shown is the average number of tokens needed to encode a document vs the number of new tokens added to the tokenizer. We compare our algorithm for tokenizer extension against \citet{yamaguchi2024can}. Impact on tokenization of (a) Wikipedia articles; (b) downstream e-commerce tasks.}
  \label{fig:tokenizer_comparison}
\end{figure*}

\subsection{Experimental setup}
\label{sec:experimental_setup}

For tokenizer training, we utilize the Hugging Face tokenizers library \citep{DBLP:journals/corr/abs-1910-03771}.
For tokenizer evaluation, we employ a comprehensive set of multilingual, e-commerce-specific, in-house downstream tasks (14 tasks in total).

As the starting LLM for tokenizer extension, we use a version of the Llama-3.1 8B model that has already been adapted towards the e-commerce domain via continuous pretraining (see \citealt{herold-etal-2025-domain} for details).
This adaptation has been done without changes to the tokenizer, so the model still uses the original Llama 3.1 tokenizer with 128k vocabulary size \cite{DBLP:journals/corr/abs-2407-21783}.

For LLM continued training, we utilize the same data as \cite{herold-etal-2025-domain}, which consists of a multilingual, 50-50  mixture of general domain and e-commerce-specific data.
As training framework, we use the Megatron-LM framework from NVIDIA \citep{DBLP:journals/corr/abs-1909-08053, DBLP:conf/sc/NarayananSCLPKV21}.
Training was conducted using 60 nodes, each having 8 NVIDIA H100 80GB GPUs (a total of 480 GPUs).
The GPUs are connected via NVIDIA NVLink (intra-node) and InfiniBand (inter-node).
Training as described in Section \ref{sec:impact_on_quality} takes less than 24h.
The hardware is part of the eBay compute platform.

\subsection{Vocabulary‑size trade‑off study}
In terms of the efficiency/speed gains from the tokenizer extension, there is a tradeoff determined by two aspects:
On the one hand, adding additional in-domain tokens should (hopefully) reduce the amount of tokens needed for the desired model output, and hence less computation in the hidden layers
of the LLM overall.
On the other hand, increasing the vocabulary size leads to more computation
(per token) in computing embeddings and logits, since the embedding/projection matrices are larger.
We have to consider this tradeoff and find an optimal point that maximizes the requests per second (RPS) of the deployed model.

Fortunately, this can be done based on an input sample and the model geometry and without expensive LLM training.
We build multiple versions of the extended tokenizer with varying amount of tokens being added, and subsequently check the average number of tokens needed to encode the text sample.
We test the encoding efficiency for a general domain dataset (English Wikipedia), as well as for our set of in-house downstream tasks.
We compare our algorithm outlined in Algorithm \ref{algo:ext_tokenizer} against the algorithm of \citet{yamaguchi2024can},
at the time of writing the only who have released their code and exact algorithm for tokenizer extension.
The results can be seen in Figure \ref{fig:tokenizer_comparison}.

As anticipated in the introduction, we find a more pronounced effect from
extending the tokenizer with e-commerce-specific tokens on the specialized domain (b)
than on general-domain Wikipedia texts (a). While Yamaguchi et al.'s approach
prepends the merge operations to the list, our approach is more conservative by appending
the new merges. This avoids the surprising decrease in efficiency when adding more tokens
seen for general-domain texts, but leads to slower gains in efficiency in specialized
text. Considering potential yet-unseen tasks, we think
that our more conservative approach with a guarantee of an upper limit on the tokenization
is preferable.

\begin{figure}[t!]
\begin{minipage}[c]{0.45\textwidth}
\vspace{0pt} % force top alignment
  \centering
  \includegraphics[width=0.99\linewidth]{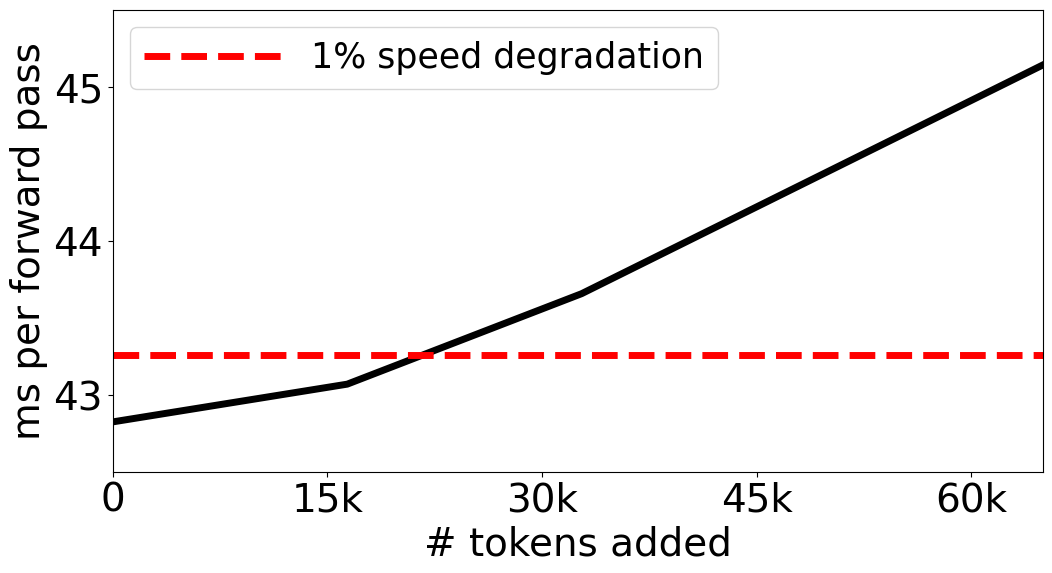} 
    \captionof{figure}{Impact of adding additional tokens to the Llama-3.1 8B model (and therefore increasing the size of embedding and projection matrices) on the speed of a single forward pass. Model is deployed via vLLM on a single H100 GPU.}
  \label{fig:forward_pass_speed}
\end{minipage}
\hfill
\begin{minipage}[c]{0.50\textwidth}
\vspace{0pt} % force top alignment
  \centering
\begin{tabular}{@{}lccc@{}}
\toprule
\multicolumn{1}{c}{Tokenizer} & \begin{tabular}[c]{@{}c@{}}\# input/output\\ words\end{tabular} & RPS & \begin{tabular}[c]{@{}c@{}}efficiency \\ gain\end{tabular} \\ \midrule
Llama & \multirow{2}{*}{300} & 29.19 & - \\
Llama ext. &  & 35.23 & 20.7\% \\ \midrule
Llama & \multirow{2}{*}{3000} & 1.95 & - \\
Llama ext. &  & 2.52 & 29.2\% \\ \bottomrule
\end{tabular}
\captionof{table}{Throughput of the Llama-3.1 8B model for a fixed number of input and output words. With the extended Llama tokenizer we need 20\% fewer tokens both for model input and output, resulting in a respective gain in RPS. Model is deployed via vLLM on a single H100 GPU.}
\label{tab:vllm}
\end{minipage}
\end{figure}

Figure  \ref{fig:forward_pass_speed} shows the effect that the increased size of
embedding and projection matrices have on the forward pass timing (8B parameters,
in vLLM framework, 300 tok/sequence at a batch size of 128 on H100).
For our 8B model, we find 30k additional tokens to be a good tradeoff between encoding and forward pass efficiency.
With 30k additional tokens, the forward pass duration increases by 1\%. Taking into account the token efficiency however,
we can expect an average speedup of 8\% on e-commerce-specific downstream tasks (see Figure \ref{fig:tokenizer_comparison}(b)), with up to 20\% speedup on specific tasks.\footnote{
We note here that the tradeoff is model size dependent, as \cite{DBLP:conf/icml/DaganSR24}
also point out: for larger models, the impact will be smaller and vice versa.}

For a more direct comparison, we measure the inference RPS of base and tokenizer-extended LLM for different input/output sequence lengths in Table \ref{tab:vllm}.
A 20\% reduction of tokens with the extended tokenizer, as we have seen for some of our downstream tasks, then shows an expected efficiency gain of around 20\% for
the shorter 300 words sequences, while for 3000 word sequences, this gain
reaches almost 30\% due to attention calculation having a larger impact.
This demonstrates that tokenizer extension is even more beneficial when dealing with longer sequences.

\subsection{Impact on model quality}
\label{sec:impact_on_quality}
Besides the efficiency tradeoff discussed above, 
we need to consider model quality, which may be affected by
the continuous training necessary for new embeddings.
Using the findings from the previous section, we extend the Llama-3 tokenizer with 30k new in-domain tokens.
Following \citet{DBLP:conf/acl/YaoHWDW21}, we initialize the new embedding and projection vectors with the average of the corresponding vectors from the existing tokens (see Algorithm \ref{algo:init_embeddings}).
We continue to train the LLM for 10,000 iterations.\footnote{Learning rate is reduced in a cosine schedule from 1.0e-5 to 5.0e-7,
all other hyperparameters are identical to \cite{herold-etal-2025-domain}.}
After training, we evaluate the LLM on the same tasks as in \cite{herold-etal-2025-domain}.
The results can be seen in Table \ref{tab:quality}.
We find that the tokenizer-extended LLM exhibits the same quality both on general domain and in-domain tasks as the domain adapted e-Llama model we started with.

\begin{table*}[]
\centering
\begin{tabular}{@{}cccccccc@{}}
\toprule
\multirow{3}{*}{Model} & \multicolumn{5}{c|}{General domain benchmarks ($\uparrow$)} & \multicolumn{2}{c}{E-commerce benchmarks ($\uparrow$)} \\ \cmidrule(l){2-8} 
 & \multicolumn{3}{c|}{En} & \multicolumn{2}{c|}{non-En} & \multicolumn{1}{c|}{En} & non-En \\
 & NLU & Lead. & \multicolumn{1}{c|}{MMLU} & NLU & \multicolumn{1}{c|}{Lead.} & \multicolumn{1}{c|}{avg.} & avg. \\ \midrule
\multicolumn{1}{l}{8B LLM} & 71.6 & 12.6 & 63.5 & 54.0 & 42.4 & 60.5 & 47.9 \\
\multicolumn{1}{l}{\quad + extend vocab} & 71.8 & 12.1 & 63.4 & 53.7 & 42.9 & 60.1 & 47.6 \\ \bottomrule
\end{tabular}
\caption {Impact of extending the Llama-3.1 model vocabulary with additional e-commerce specific tokens on the model quality.}
\label{tab:quality}
\end{table*}

\subsection{Behavior of Tokenizer-Extended LLM}
\label{subsec:behavior}

In this section we discuss how well the tokenizer-extended LLM actually utilizes the newly added in-domain tokens.
In theory, the model could simply ignore all new tokens during generation, and we do not get any benefit of the tokenizer extension.
We note that this only applies to the text generation.
The input that the model receives at the beginning will of course always be tokenized using the new tokens, so here we are guaranteed to get the benefit of more efficient encoding.
To the best of our knowledge, we are the first to analyze this crucial part of LLM tokenizer extension.

We take again our set of in-house downstream tasks.
For each example, we go through the text word-by-word.
For each word, we look at the probability distribution of the tokenizer-extended LLM model, given all previous words as context.
Going through the words, we count, how often the model wants to predict the \lq{}old\rq{} tokenized version of the word, and how often it wants to predict the \lq{}new\rq{} tokenized version.
The results are shown in Table \ref{tab:token_usage}.

\begin{table}[]
\centering
\begin{tabular}{@{}cc@{}}
\toprule
\begin{tabular}[c]{@{}c@{}}sequence length\\ {[}\# words{]}\end{tabular} & New tokenization being used \\ \midrule
\textless 15 & 95.3\% \\
15 - 49 & 98.0\% \\
\textgreater{}= 50 & 97.8\% \\ \bottomrule
\end{tabular}
\caption {How frequently is the new tokenization used vs the old tokenization. For sequences larger than 15 words, the adapted model prefers the new tokenization in roughly 98\% of cases.}
\label{tab:token_usage}
\end{table}

We find, that in almost all cases, the model prefers the new tokenization vs the old one.
For short sequences \textless 15 words, there is still around 5\% chance that the model produces the old tokenization, but for longer sequences, the adapted model prefers the new tokenization in almost 100\% of cases.

\section{Conclusion}

In this work we identify vocabulary mismatch as an often‑overlooked, yet decisive, bottleneck when autoregressive LLMs are moved from broad‑domain pretraining to the specialized distribution of a specific target domain.
We present a tokenizer‑extension algorithm that adds the most frequent in‑domain tokens without ever increasing the number of tokens required to encode any sequence, thus guaranteeing backward compatibility.

Experiments on a multilingual, production‑grade e‑commerce set of in-house downstream tasks show that the extended tokenizer shortens input sequences by up to 20\%.
When deployed, this translates into 20\%-30\% higher throughput, depending on prompt length. 
Importantly, this is achieved without compromising model quality on general or in-domain tasks.
In additional studies, we find that the model learns to emit the new tokens in $\approx$98\% of cases for sequences longer than 15 words, confirming that the additional vocabulary is effectively embraced rather than ignored.
\bibliographystyle{plainnat}

\bibliography{neurips_2025}

%%%%%%%%%%%%%%%%%%%%%%%%%%%%%%%%%%%%%%%%%%%%%%%%%%%%%%%%%%%%

\appendix

\section{Technical Appendices and Supplementary Material}
% Technical appendices with additional results, figures, graphs and proofs may be submitted with the paper submission before the full submission deadline (see above), or as a separate PDF in the ZIP file below before the supplementary material deadline. There is no page limit for the technical appendices.

\subsection{Limitations}
\label{sec:limitations}
In this study we focus on a single domain, namely e-commerce, since this is the domain of our in-house downstream tasks.
Potentially, the behavior of our approach might be different for different domains.
Furthermore, we focus on a single model family, namely the Llama-3 models, since they are the most relevant for us at this time.
The behavior of other model families might be different in terms of potential savings or the change in model quality after the tokenizer extension.

\subsection{Related Work - Continuation}
\label{sec:related_work_cont}

\subsubsection*{LLM Efficiency}

Efficient LLM inference can leverage a spectrum of model compression and system-level optimizations. 
Model quantization reduces weight/activation precision (e.g., 8-bit or 4-bit) to shrink memory and computation, achieving significant throughput gains with negligible accuracy loss \cite{DBLP:conf/icml/XiaoLSWDH23}.
Similarly, model pruning eliminates redundant parameters or entire sub-modules (e.g., attention heads or layers), yielding substantially smaller and faster models with minimal performance drop \cite{DBLP:conf/emnlp/LagunasCSR21}.
Knowledge distillation trains compact student models to mimic large teacher models, retaining most of the original accuracy with far fewer parameters and faster runtime \cite{DBLP:journals/corr/abs-1910-01108}. 
Beyond model-centric methods, compilation frameworks (e.g., NVIDIA TensorRT and Apache TVM) convert neural network graphs into optimized executables via kernel fusion, mixed-precision and other low-level optimizations, greatly improving throughput on target hardware \cite{electronics14152977}. 
Finally, edge deployment of LLMs has emerged, using aggressive compression and efficient runtimes to run models on resource-constrained devices \cite{DBLP:journals/corr/abs-2504-03360}.

Our approach of vocabulary extension is fully orthogonal to existing efficiency-related optimizations.
It can be used on top of any highly efficient deployment to further reduce cost and latency.

\subsubsection*{Domain Adaptation}

Large pretrained LLMs often struggle with domain-specific tasks, motivating targeted domain adaptation \cite{DBLP:conf/nips/LewkowyczADDMRS22, DBLP:journals/corr/abs-2311-16079, DBLP:journals/corr/abs-2308-12950}.
Continuing pretraining on in-domain text or fine-tuning the entire model can substantially boost performance on domain-specific tasks \cite{DBLP:conf/iclr/AzerbayevSPSMJD24, DBLP:journals/corr/abs-2402-03300, DBLP:journals/corr/abs-2401-09646}, but performing these is quite expensive.
To improve efficiency, parameter-efficient methods have emerged: adapter-based fine-tuning inserts small trainable modules into the network while keeping most weights frozen \cite{DBLP:conf/icml/HoulsbyGJMLGAG19}, and prompt tuning optimizes a few soft prompt vectors with the model fixed \cite{DBLP:conf/emnlp/LesterAC21}.
Retrieval-based approaches further inject relevant domain knowledge at inference time, enabling LLMs to leverage external in-domain data on the fly instead of exhaustive retraining \cite{xu-etal-2023-retrieval}.

\subsection{Detailed Methodology}
\label{sec:detailed_,ethodology}

\subsubsection{Training a Domain-Specific Tokenizer}

The first step involves the training of a new in-domain tokenizer model on some domain-specific dataset. 
It is important that we choose sufficient training data representative of the target domain, ideally encompassing all potential downstream tasks.
For the tokenizer, we only care about the vocabulary, we can discard the list of merge operations.
Since we expect a significant vocabulary overlap between the original tokenizer and the in-domain one, we should set the target token size for training to a higher number than the number of tokens we want to add to the original tokenizer.

\subsubsection{Extending the Original Tokenizer}

We initialize the extended tokenizer with the original tokenizer.
We then traverse the vocabulary of the in-domain tokenizer in the default order.
For each token, we tokenize it using the current extended tokenizer.
If the result is a single token, this means this token is already in the vocabulary of the extended tokenizer and we can continue without doing anything.
However, if the token is split into two tokens, we append the tuple of the two tokens at the end of the list of merge operations of the extended tokenizer and also add the token into the model vocabulary.
We want to point out, that following this approach, we will never encounter the case where the token is split into three or more tokens, because all except one of the corresponding merge operations will already have been added to the extended tokenizer in an earlier step. 
We repeat the above until we have added a fixed amount of new tokens to the tokenizer.
The detailed algorithm is outlined in Algorithm \ref{algo:ext_tokenizer}.

Since we append the new merge operations always to the end of the list, the first part of the final list of merge operations will be identical to that of the original tokenizer.
The merges that are appended to the list (see Algorithm \ref{algo:ext_tokenizer} line 14) can only reduce (or leave unchanged) token counts because earlier merges are unaffected by later ones. 
That means, contrary to existing approaches, we will never have the situation where our extended tokenizer results in a worse tokenization than the original one.

\subsubsection{Embedding Initialization}

Since we are extending the LLM vocabulary, we also need to initialize the newly added embedding and projection vectors in the LLM.
The baseline method comprises of using random initialization, but previous work has shown that better performance can be achieved when initializing the new token-embedding as the average of the embeddings of the existing tokens that the new token is composed of \cite{DBLP:conf/acl/YaoHWDW21}.
Therefore we follow this approach of average initialization.
The detailed algorithm for embedding initialization is outlined in Algorithm \ref{algo:init_embeddings}.

\subsubsection{Continuous Training}

To adapt the tokenizer-extended LLM to the new extended vocabulary, we need to continuously train the model on some data.
In our setting, we start with a model that is already adapted towards the e-commerce domain, so we just sample training data from the same distribution that was used to adapt the model to the domain in the first place.
We perform full training of all model parameters, not freezing any part of the network.

\subsubsection{Final Evaluation}

We need to evaluate the final tokenizer-extended LLM in several aspects:
\begin{itemize}
    \item What is the inference speedup on in-domain downstream tasks? 
    \item How has the model quality changed?
    \item To what extent is the model utilizing the new tokenizations vs the old one?
\end{itemize}
To measure the model quality, we utilize the same set of public and e-commerce specific benchmarks as in \cite{herold-etal-2025-domain}.
To measure the inference speed and utilization of new tokens, we design a set of experiments utilizing a comprehensive set of in-house downstream tasks and LLM deployment via vLLM \cite{DBLP:conf/sosp/KwonLZ0ZY0ZS23}.

\begin{algorithm}[t]
\caption{Extend base BPE Tokenizer with in-domain tokens}
\label{algo:ext_tokenizer}
\KwData{$T_b$: base tokenizer, $T_d$: in-domain tokenizer, $N$: number of tokens to add}
\KwResult{$T_{\text{ext}}$: extended tokenizer}

$V_b \gets$ vocab of $T_b$\;
$M_b \gets$ merges of $T_b$\;
$V_d \gets$ vocab of $T_d$, sorted by descending frequency\;
$i \gets 0$\;

\ForEach{$t \in V_d$}{
    \If{$t \in V_b$ \textbf{or} $T_b$'s pre-tokenizer splits $t$}{
        \textbf{continue}\;
    }
    $enc \gets T_b.\text{encode}(t)$\;
    \If{$\text{len}(enc) \ne 2$}{
        \textbf{continue}\;
    }
    % $[l, r] \gets enc$\;
    $[l, r] \gets T_b.\text{encode}(t)$\;
    $m \gets$ merge of $l$ and $r$\;
    \If{$m \notin M_b$}{
        append $m$ to $M_b$\;
    }
    add $t$ to $V_b$ with new ID\;
    $i \gets i + 1$\;
    \If{$i \ge N$}{
        \textbf{break}\;
    }
}
$T_{ext} \gets T_{b}$\;
\Return updated tokenizer $T_{\text{ext}}$\;
\end{algorithm}

\begin{algorithm}[t]
\caption{Initialize the in-domain token embeddings.}
\label{algo:init_embeddings}
\KwData{$T_b$: base tokenizer, $T_d$: in-domain tokenizer,  $E_b$: base token embeddings}
\KwResult{$E_d$: in-domain token embeddings}

$V_b \gets$ vocab of $T_b$\;
$V_d \gets$ vocab of $T_d$\;
$E_d \gets$ \{\}\;

\ForEach{$t \in V_d$}{
    \If{$t \in V_b$}{
        \textbf{continue}\;
    }
    $e \gets 0$\;
    $t_b \gets T_b.\text{encode}(t)$\;
    \ForEach{$t_{old} \in t_b$}{
    $e \gets e + E_b[t_{old}]$\;
    }
    
    $e \gets e / \text{len}(t_b)$\;
    add $e$ to $E_d$\;
}
\Return $E_d$\;
\end{algorithm}

\end{document}